\documentclass[preprint,12pt]{elsarticle}

\usepackage{amssymb}
\usepackage{amsmath}
\usepackage{wrapfig}
\usepackage{float}
\usepackage{tabularx}
\usepackage{longtable} 
\usepackage{array} 
\usepackage{hyperref}

\journal{Computer Speech and Language}

\begin{document}
\sloppy

\begin{frontmatter}



\title{BERSting at the Screams: A Benchmark for Distanced, Emotional and Shouted Speech Recognition.}

\author[label1,label2,label3]{Paige Tutt\"os\'i}
\author[label1]{Mantaj Dhillon}
\author[label1]{Luna Sang}
\author[label1]{Shane Eastwood}
\author[label1]{Poorvi Bhatia}
\author[label1]{Quang Minh Dinh}
\author[label1]{Avni Kapoor}
\author[label1]{Yewon Jin}
\author[label1]{Angelica Lim}
\affiliation[label1]{organization={School of Computing Science, Simon Fraser University},
             addressline={8888 University Drive},
             city={Burnaby},
             postcode={V5A 1S6},
             state={British Columbia},
             country={Canada}}

\affiliation[label2]{organization={SUPMICROTECH, CNRS, Institut FEMTO-ST, Université de Franche-Comté},
             addressline={24 rue Alain Savary},
             city={Besançon},
             postcode={25000},
             country={France}}
\affiliation[label3]{organization={Enchanted Tools},
             addressline={18 Rue de la Fontaine au Roi},
             city={Pairs},
             postcode={25000},
             country={France}}



\begin{abstract}
Some speech recognition tasks, such as automatic speech recognition (ASR), are approaching or have reached human performance in many reported metrics. Yet, they continue to struggle in complex, real-world, situations, such as with distanced speech. Previous challenges have released datasets to address the issue of distanced ASR, however, the focus remains primarily on distance, specifically relying on multi-microphone array systems. Here we present the B(asic) E(motion) R(andom phrase) S(hou)t(s) (BERSt) dataset. The dataset contains almost 4 hours of English speech from 98 actors with varying regional and non-native accents. The data was collected on smartphones in the actors homes and therefore includes at least 98 different acoustic environments. The data also includes 7 different emotion prompts and both shouted and spoken utterances. The smartphones were places in 19 different positions, including obstructions and being in a different room than the actor. This data is publicly available for use and can be used to evaluate a variety of speech recognition tasks, including: ASR, shout detection, and speech emotion recognition (SER). We provide initial benchmarks for ASR and SER tasks, and find that ASR degrades both with an increase in distance and shout level and shows varied performance depending on the intended emotion. Our results show that the BERSt dataset is challenging for both ASR and SER tasks and continued work is needed to improve the robustness of such systems for more accurate real-world use.
\end{abstract}


\begin{highlights}
\item A dataset of distanced, emotional speech, spoken and shouted recorded on smart phones
\item Speech features differ for different distances, shout levels and emotions
\item State of the art automatic speech recognition performance degrades with distance
\item Speech emotion recognition performs poorly regardless of distance
\end{highlights}


\end{frontmatter}



\section{Introduction}

One day we hope to be able to call out to our phone in another room when frustrated about a task and still have it understand our every word. Current state-of-the-art for automatic speech recognition (ASR) continues to struggle with distanced, shouted, and emotional speech. This is not only a problem of convenience, but also of safety and security. In emergency situations, we may not be able to move closer to a microphone or may not be able to use a calm voice. It is exactly in these types of situations that we must have reliable speech recognition for distanced, high arousal speech.

One of the primary problems facing advancement in this area is a lack of data available for training and benchmarking of ASR in distant and expressive contexts. Traditionally, many datasets used for these tasks are professionally recorded, close to the microphone, and without emotions. Publicly available data, such as Youtube videos \cite{youtube} and podcasts \cite{Lotfian_2019_3}, are also used, however, this type of data lacks annotations, specifically those pertaining to microphone distance, that may help improve the development of systems robust to expressive, distanced, and high-intensity speech. Curated datasets have also focused on far field, multi-microphone, data collection \cite{cornell24_chime}, which is not always available to ASR systems in the wild.

To address this lack of data, we introduce BERSt\footnote{https://huggingface.co/datasets/chocobearz/BERSt}: a dataset with emotional, distanced, and high arousal speech recorded on smartphones in various home environments. BERSt contains 4526 audio clips (approximately 3.75 hours) of data from 98 actors in 98 households around the world. This dataset is intended to help fine-tune and benchmark speech recognition tasks such as automatic speech recognition, shout recognition and speech emotion recognition (SER). The dataset contains distanced, emotionally acted speech with varying levels of arousal. Here we present:
\begin{enumerate}
\item An introduction of the data collection methods, actor demographics, and available ground truth annotations.
\item Exploration of the acoustic, vocal features of speech at varying distances and for different emotions.
\item Out-of-box benchmarks of current state-of-the-art ASR and SER models.
\end{enumerate}

\section{Related Work}
As the BERSt dataset contains distanced, emotional, and shouted speech that can be used for multiple speech recognition tasks, we present the related work as follows: ASR for Emotional Speech, far-field ASR and datasets, far-field SER and  datasets.

\subsection{Automatic Speech Recognition}

\subsubsection{ASR for emotional speech}
The disruption of emotions on the performance of ASR is a known problem, as most ASR systems are trained on neutral speech. Yet, limited research has looked at how to improve the robustness of ASR for emotional speech. \cite{10.1007/s10772-018-9503-z} and \cite{Kammili2021HandlingES} investigated Telugu language ASR and found that performance degraded greatly for emotional speech. They then trained their model on an augmented dataset, where they modified the F0 (pitch), energy, and speaking rate to add variation similar to emotional speech and used this as training data for their model. In \cite{10.1007/s11036-018-1052-9}, SER was used as a primary step to select emotionally specific ASR models. In \cite{santoso21_interspeech}, the authors created a confidence measure that adjusts the weights in the ASR depending on the likelihood of an SER error in each word. Some research has looked at why emotional speech poses a problem for ASR systems. \cite{ATHANASELIS2005437} suggests that vocal features, such as: source, intensity, speech quality, prosody, and timing, effect ASR performance. However, as suggested by \cite{li2023asremotionalspeechwordlevel}, the emotional text itself may pose difficulties for ASR systems. They also found that when benchmarking models on SER datasets, the neutral emotions may not always have the best performance, and postulate that this may be due to neutral utterances being, in general, shorter. Additionally, there is work that looks primarily at SER but includes ASR in the recognition pipeline \cite{9746289, sahu19_interspeech, 4960651}. Our dataset provides transcriptions for emotional speech that can be used to help benchmark the performance of ASR models for different intended emotions. There currently exists no dataset specific to emotional speech for ASR.

\subsubsection{Far-field ASR}
Distanced ASR has more extensive coverage in the literature, mostly driven by the CHiME challenge \cite{MA2013820, 6707723, 7404837, chen18d_interspeech, barker18_interspeech, watanabe2020chime, cornell23_chime, cornell24_chime}. It has been acknowledged \cite{10.1145/3335595.3335635} that ASR shows a marked deterioration in model performance as distance increases. Most distant speech models and datasets rely on microphone arrays, which are not present in our dataset, and may pose a particular challenge, but also reflect lower resource, i.e. single microphone, in-the-wild environments, such as someone in their home speaking to their voice assistant on their phone. Early work in distanced ASR investigated the combination of multichannel signal processing and neural networks \cite{6843274}, as well as long short-term memory (LSTM) recurrent neural networks (RNNs) \cite{10.1109/ICASSP.2016.7472780, 8326195}, and deep neural networks (DNNs) \cite{DBLP:journals/corr/abs-1712-06086}. The VOiCES from a Distance Challenge \cite{nandwana19b_interspeech} focused instead on single channel ASR. More recent work has looked at quaternion neural networks \cite{qiu20_interspeech}, but the majority of the research has been focused on leveraging other audio techniques, often related to microphone arrays or speaker diarization \cite{bando24_interspeech} to improve distanced ASR. \cite{cornell21_interspeech} for example, created a ranking framework to select the best channel for speech recognition. In \cite{9189820}, the authors used a combination of data processing with dereverberation, source separation, and acoustic beamforming combined with an ASR that had multicondition training and adaptation. The BERSt dataset contains annotated distances, as well as obstructions to the microphone for a singular, smartphone microphone, and provides a challenging new benchmark for distanced ASR systems.

\emph{Far-field ASR datasets}:
Many large datasets exist for ASR training, fine-tuning, and evaluation, and more recently, datasets contain spontaneous speech that can include emotional and shouted speech in some cases. Yet, emotion is not a feature of these datasets and annotations for these features are not provided. Some datasets include low quality audio recordings from telephones (although not necessarily cell phones) \cite{cieri-etal-2004-fisher, switchboard, cornell24_chime}. The SUSAS (Speech Under Simulated and Actual Stress) dataset \cite{hansen97b_eurospeech} explicitly contains elicited stressed and emotional speech yet does not contain different distances for far-field ASR tasks. Multiple datasets contain environmentally noisy and reverberant data \cite{oneill21_INTERSPEECH, cornell24_chime, richey2018voices}.

The CHiME Challenge \cite{cornell24_chime} and VOiCES \cite{richey2018voices} datasets are both specifically designed for distanced ASR tasks. The CHiME-6 and DiPCo datasets contain recordings of dinner parties between 4 participants across different rooms in a home. It contains conversational speech and a high amount of environmental noise. Mixer 6 consists of dyad interview pairs and includes a prompt reading and a telephone recording. NOTSOFAR-1 contains audio data of 30 different meetings, between 32 participants (up to 8 speakers at a time). It is likely that some emotional speech exists in these CHiME datasets, but this is not explicitly elicited, especially for high arousal shouted data. Moreover, all the recordings include multi-microphone scenarios, but, in some cases, such as for NOTSOFAR-1, cell phones are used as the recording devices. The VOiCES dataset has 4 different room configuration containing different microphones, noise distractors, and loudspeaker angles. The data was played over loudspeakers from the Librispeech \cite{libspeech} Corpus, which does not contain emotional nor shouted speech. BERSt expands speech recognition datasets by adding emotional and shouted speech. There are few phrases, but they remain phonetically diverse. Additionally, none of the distanced ASR datasets contain as many different home environments as BERSt, which contains at least 98 different rooms. Moreover, BERSt has not only different distances, but also different surfaces and obstructions to the microphone that are not present in the other datasets. Finally, BERSt contains a range of regional and non-native accents, which can help improve fairness in ASR systems.

\subsection{Far-field SER}\label{serdist}
Distant SER is relativity less well explored than distanced ASR, yet, there has been work in this domain. \cite{10.1145/3144457.3144503} trained their model with added reverberation and applied preprocessing on the audio to tidy the signal before classification. In \cite{10.1145/3130961}, the authors used a dataset with synthetic distance modifications to train an LSTM. They also explored the dataset to extract the features that are most robust to changes in distance, in order to be used for classification. \cite{10.1007/978-3-031-23804-8_15} also used a dataset overlaid with synthetic reverberation and room noise and found that a deep convolutional neural network trained on a combination of noised and clean data had comparable performance on noisy and clean data. \cite{grageda23_interspeech} and \cite{garcia24_interspeech} used beamforming and room impulse response to help classify distant emotions in a human robot interaction scenario. They did not have a dataset of in-situ produced distanced speech, but rather used a general SER dataset, MSP-Podcast \cite{Lotfian_2019_3}, played through speakers at varying distances and angles.

\emph{Far-field SER datasets}:
As our data has not been validated for emotion labels, we do not claim that it is currently a dataset that can be used to train SER models. However, it can be used as an evaluation tool of SER systems for distanced and shouted emotional speech based on the actor's intended expressions. As was seen in Section \ref{serdist}, most of the work with distanced SER has been done with synthetic addition of distanced noise.

\section{Dataset}

\subsection{Collection}
\begin{figure}[]
  \centering
  \includegraphics[width=\linewidth]{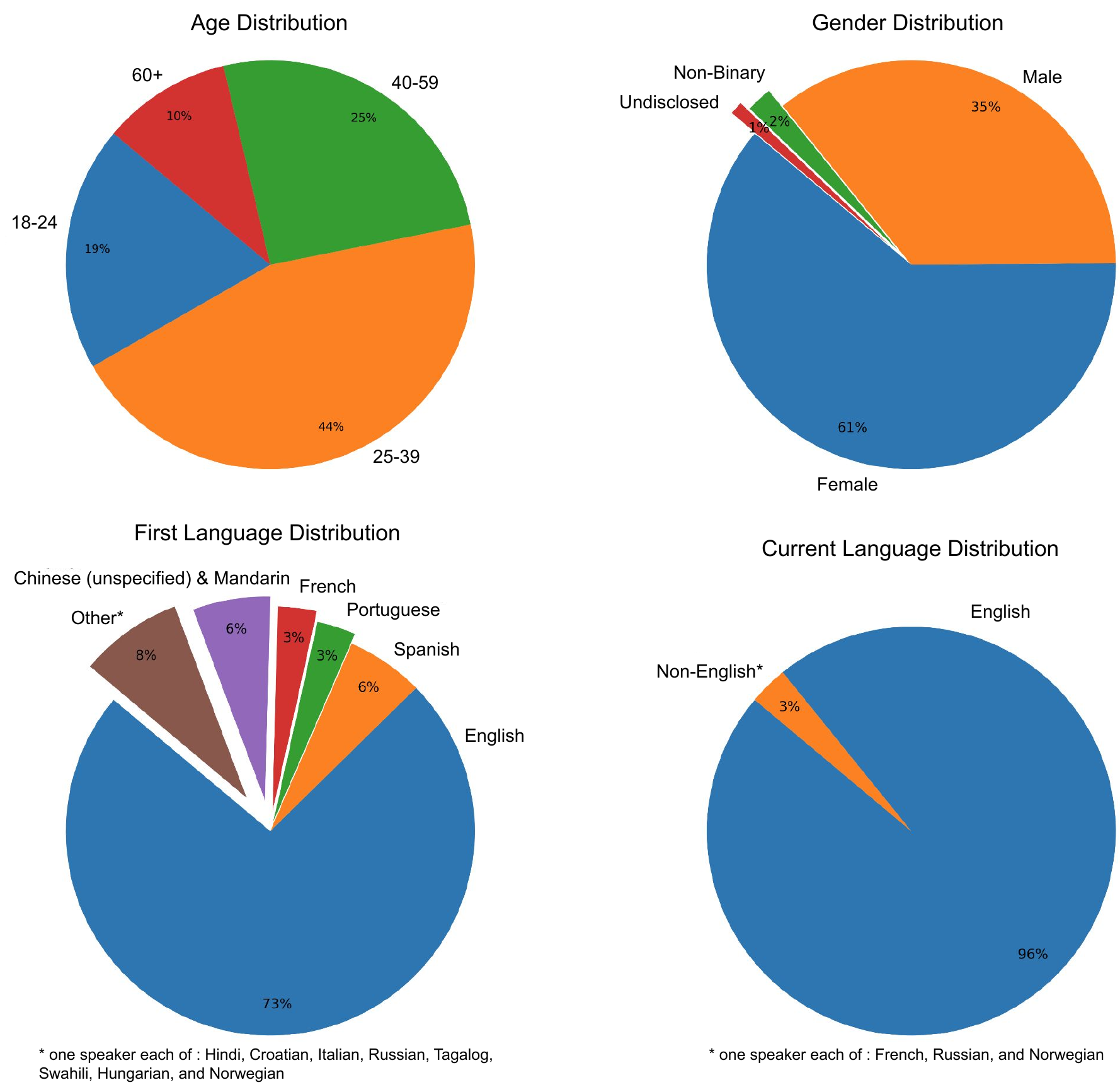}
  \caption{Collected participant demographic information.}
  \label{dem}
\end{figure}

We collected the B(asic) E(motion) R(andom phrase) S(hou)t(s) (BERSt) dataset. Ninety-eight professional actors were recruited through Backstage Casting\footnote{https://www.backstage.com/}. The data was collected entirely remotely, allowing the actors to work on their own time, in their own space. This also means that the data contains at least 98 different rooms. The actors were specifically told not to worry about minimizing background noise to allow for more natural environments. To facilitate this flexible means of collection, a smartphone application was built; example pages from the application can be see in the Appendix (Figure \ref{app}). The application first collected demographic information about the actor, see Figure. \ref{dem}. The primary current language, i.e. the language most spoken daily, for the actors was English (N=95), however, regional accents included varying parts of Canada, several states in the USA, the UK, Ireland, and Australia. Additional non-native accents are present in the dataset (N=26), as other first languages included: Spanish, Portuguese, French, Chinese, Hindi, Croatian, Italian, Russian, Tagalog, Swahili, Hungarian, and Norwegian. Including various accents of native English speakers, as well as foreign accents is important for developing robust speech recognition systems. \cite{FENG2024101567} found that modern ASR systems still present strong performance biases for both regional and non-native accents.

\begin{table}[t]
  \centering
    \centering
    \scriptsize
    \begin{tabular}{| c |}
      \hline
      {\textbf{Phrase}} \\
      \hline
      banana and mustard sandwiches~~~ \\
      hungry action hippos fruit~~~ \\
      another dog secretary show~~~ \\
      snow meteor down the chimney~~~\\
      birds make new jingles~~~\\
      heavy undersea birthday pumpkins~~~\\
      fluffy baseboard yogurt division~~~\\
      zebra cats walking on lamps~~~\\
      five special guest treasures~~~\\
      ripped ocean jumper~~~\\
      five dollar slaw onion cure~~~\\
      the thunder break craze~~~\\
      join the ear look club~~~ \\
      wolf ring lights are fantastic~~~ \\
      \hline
    \end{tabular}
    \caption{Phrases for data collection.}
    \label{phrase}
\end{table}

Next, the actors were guided through a series of recordings where they were given a random phrase, emotion, and phone position. The 14 phrases can be seen in Table \ref{phrase}. These phrases were chosen to be nonsensical, yet to cover the range of English phonemes. In addition, an attempt was made to not have any sentences with particularly strong emotions, ``Wolf ring lights are fantastic," however, is an exception, displaying a positive sentiment. The use of nonsense phrases poses a particular challenge for speech recognition systems, as they are not able to rely on linguistic context to predict the words or emotions, and most of the phrase has high surprisal, i.e. words are easier to comprehend in contexts where they are highly predictable, if not they have higher surprisal \cite{LEVY20081126}. For each recording, the actors were instructed to ``speak the phrase", ``shout the phrase", and ``scream the phrase". This creates varying intensity levels within the emotions, which poses a challenge for speech recognition models that may rely heavily on intensity, for example, when predicting arousal levels. The combination of both distanced recordings and varying speaking volumes means that intensity alone cannot define neither distance, nor arousal level.

\begin{figure}[]
  \centering
  \includegraphics[width=\linewidth]{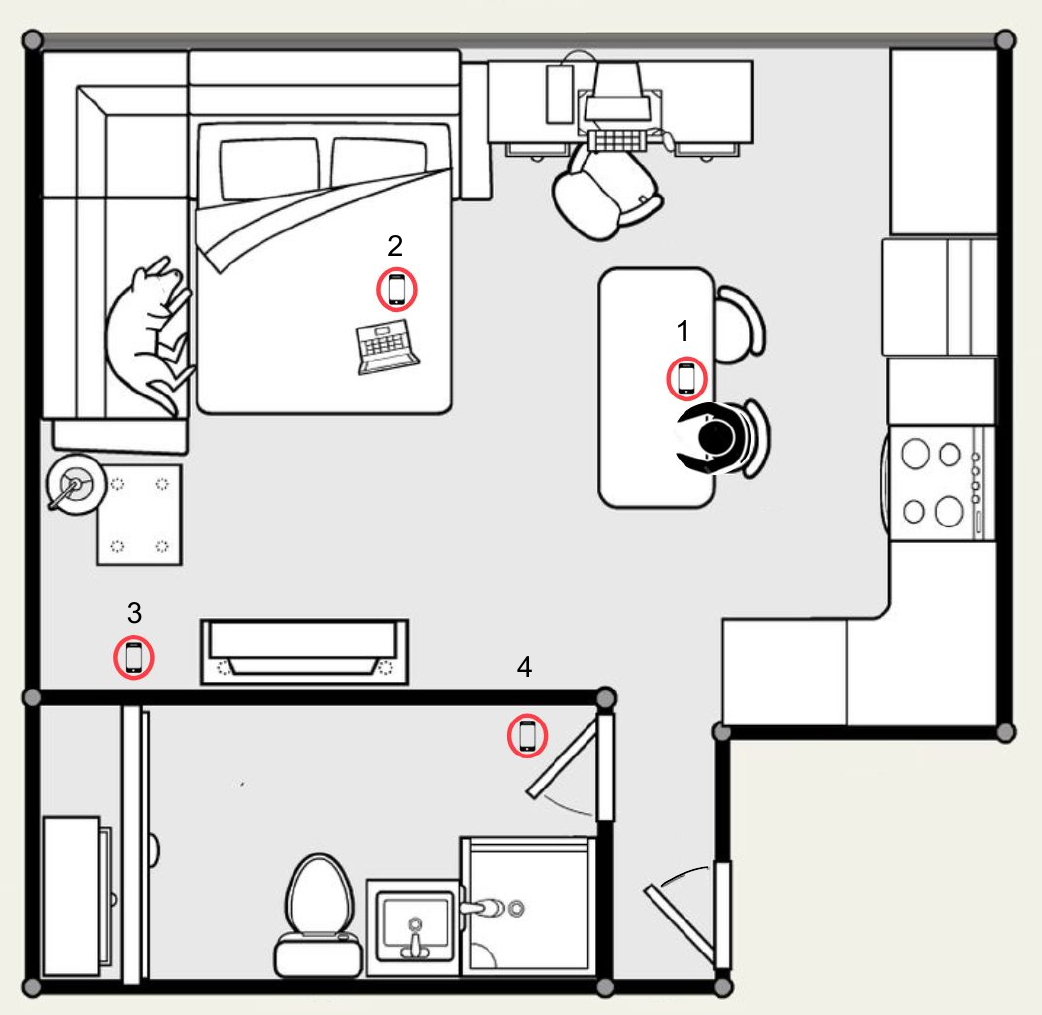}
  \caption{Phone positions for data collection. Positions include 4 distance levels: 1. Near the body, 2. 1-2m distance, 3. Opposite side of the room, and 4. Outside the room.}
  \label{positionsimg}
\end{figure}

The actors completed 19 recordings progressively moving the phone further away; the phone positions can be seen in Table \ref{positiontab} and Figure \ref{positionsimg}. Exact measurements were not taken, rather the actors were asked to use their best approximations.  Not only was the phone moved further from the speaker, but, in the furthest location, it was actually placed on the other side of a wall or door. Additional obstructions were included at each distance, including putting the phone in a bag, putting the phone in their pocket, or covering the microphone with their hand. Lastly, the phone was placed on both soft and hard surfaces to capture any possible differences in audio recording quality that this may cause \cite{rust2021, Dupont2020CharacterizationOA}.

\begin{table}[t]
  \centering
  \resizebox{\textwidth}{!}{
  \begin{tabular}{| l | l | }
    \hline
    {\textbf{Phone distance}} & {\textbf{Positions}}\\
    \hline
    next to your face  & mic facing your mouth as you would in a phone conversation,  ~~~\\
    & mic/phone facing away from your face, hand covering the mic~~~\\
    next to your hip &  back of phone on your palm, hand covering the mic ~~~\\
    in your pocket & N/A~~~\\
    in a bag and hold it next to your hip & N/A~~~\\
    1-2 meters away &  face up on any surface, face down on a hard surface, ~~~\\
    & face down on a soft surface, in a bag~~~\\
    on the opposite side of the room & face up on any surface, face down on a hard surface, ~~~\\
    & face down on a soft surface, in a bag~~~\\
    as far away as possible, on the opposite side of a wall & face up on any surface, face down on a hard surface, ~~~\\
    & face down on a soft surface, in a bag~~~\\
    \hline
  \end{tabular}}
    \caption{Phone positions for data collection.}
    \label{positiontab}
\end{table}

Seven basic emotions were prompted: \emph{neutral}, \emph{anger}, \emph{joy}, \emph{fear}, \emph{disgust}, \emph{surprise}, and \emph{sadness}. The actors were prompted with basic emotions for two reasons: firstly, basic emotions are non-specifc and leave interpretations, e.g. hot or cool anger, up to the actors. Therefore, by prompting basic emotions, we are able to ensure we capture a range of emotions that are available for researchers to label for their own use cases. Secondly, having the basic emotion ground truth labels opens up the dataset for use in more generalized SER tasks, which are often limited to basic emotions (in many cases only 4: neutral, sadness, joy and anger). Altogether, each actor recorded 19 phrases, at 19 locations and three loudness levels resulting in 57 recordings per actor prior to data cleaning.

\subsection{Data cleaning}
The actors made one single recording for each phrase, emotion, and phone position. This means that each raw audio clip contains three utterances, one per intensity level. To split these, we used a voice activity detection (VAD) toolkit \cite{VAD}. To confirm that all the data was correctly split, a native English speaking annotator listened to all of the utterances and flagged those that either had not been correctly split, or where the actor made a mistake in their phrase and the transcription needed to be updated. From this, 4 native English speaking annotators split the remaining data by hand using Audacity\footnote{https://www.audacityteam.org/} and corrected the transcriptions to reflect what the actor was truly saying, for example changing ``ripped ocean jumper" to ``a ripped open jumper". Additionally, clips where there was only noise, e.g. the phone was too far away to capture any discernible speech, were removed.

To then annotate the shout levels, two annotators labeled the data as ``shouting" or ``not-shouting". An attempt was made to assign the original three shout levels: speaking, shouting, and screaming. However, an early pilot of these annotations using data scraped from Reddit (16 samples), YouTube (6 samples), and Streamable (1 sample), found that annotators had difficulty discerning three levels of intensity, but could easily label two. Twenty-five percent of the data was labeled by these two annotators and a Cohen's kappa score was calculated to assess annotator agreement. The Kappa score was 0.80 (substantial agreement), and as such a single annotator was used to label the remaining data, and their labels are used as the definitive labels provided with the dataset. Emotion validation has not been completed and remains future work.

We provide a train, test, and validation split for the data at a 80/10/10 ratio. We have no speaker crossover between the splits, and therefore attempted to maintain a balance of distances, shout levels, and emotions across the sets.

\section{Data exploration} \label{explore}
\begin{figure}
  \centering
  \includegraphics[width=\textwidth]{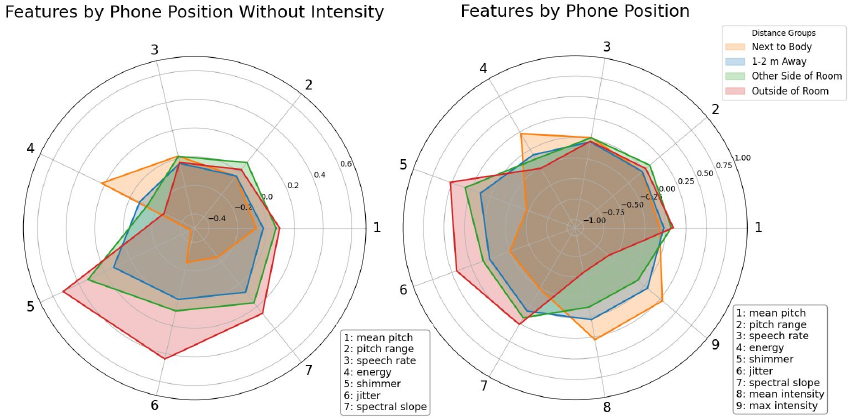}
  \caption{Extracted vocal features for each phone position (normalized), without (left) and with (right) intensity features.}
  \label{distall}
\end{figure} 

To explore the audio and vocal features of the data we used the openSMILE python toolbox from AudEERING \cite{opensmile}. We extracted the complete Geneva Minimalistic Acoustic Parameter Set (GeMAPS) and attempted to explore the data using principal component analysis and t-SNE\footnote{https://scikit-learn.org/1.5/modules/generated/sklearn.manifold.TSNE.html}, yet, we found we did not find any useful information in this dense data. Instead we opted to extract a set of features as used in \cite{readroom} using the associated voice toolbox\footnote{https://github.com/ehughson/voice\_toolbox}. These features include: 1) Loudness Features: (a) mean intensity, (b) energy, (c) maximum intensity. 2) Spectral Features: (a) mean pitch, (b) pitch range, (c) shimmer, (d) jitter, and (e) spectral slope. 3) Rate-of-Speech Features: syllables per second. The features were normalized across the dataset, yet, we found that, due to the shouted data having very high intensity, in combinations with the distances, reducing the intensity, these features were overpowering on the plots when using a similar axes scaling across plots. Therefore, we created visualizations both with and without intensity features, to better understand more subtle differences not only controlled by intensity levels. If no plot is shown for a given feature set containing intensity levels, this means that the intensity for all groups was similar (Figure \ref{emoall}).

 We explore the primary elements of the dataset: distance, shouts, and emotions, and how they interact with one another. This analysis provides important information on how vocal features differ for the different elements in the data set and may provide suggestions of features that are confounding to include as primary features in far-field, shouted, and emotional audio models.

\subsection{Distance Features}
First we explore how audio features differ per distance, over all audio samples (Figure \ref{distall}). As expected, we observe that both mean and max intensity decrease as the phone is moved away from the body, especially when the phone is in another room (right). We also observe higher energy levels when near the body, and higher shimmer and jitter (typically associated with vocal fry and hoarseness, respectively) for recordings at a distance. We note especially low values for shimmer, jitter, and spectral slope when near the body. 

\begin{figure}
  \centering
  \includegraphics[width=\textwidth]{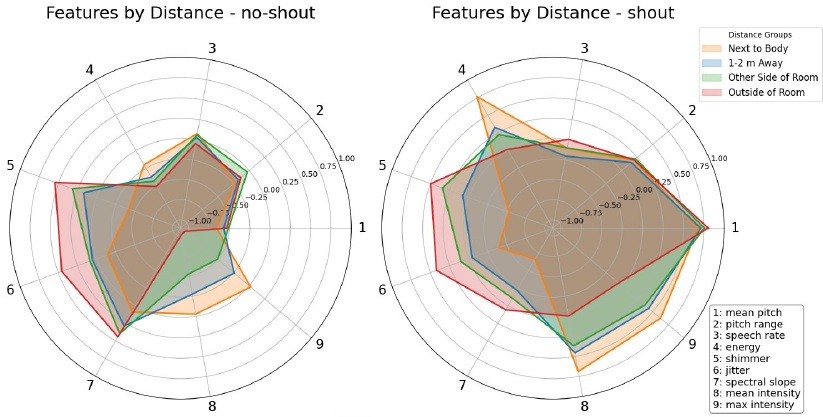}
  \caption{Extracted vocal features for each phone position (normalized), for non shouted (left) and shouted (right) samples.}
  \label{distshout}
\end{figure}

We next inspected the distance features for shouted and non-shouted speech (Figure \ref{distshout}). We see that the vocal features for distance groups show a marked difference depending on whether the data is shouted, or not shouted. For both shouted and not shouted data we see the same increase in intensity as the phone is moved closer to the body, however, we see that the pitch is much higher for shouted than for not shouted data at all phone positions. Additionally, we see a marked increase in energy for samples recorded close to the body when shouted. Differences in shimmer and jitter seem to be driven more by the phone location than by the shout level.

These features provide insights into why distant speech recognition tasks remain challenging. We see that the distance from the microphone is an important factor in determining the audio features of a speech sample. 
\textbf{Specifically, alongside the expected differences in intensity (the change in energy over a given area) and total energy transferred, shimmer and jitter are primarily controlled by the distance from the microphone. Both of these features diminish as the microphone moves away from the body. This may be a result of these features being more difficult to extract from distant audio (similar to details being lost in a photo being taken from afar) and that a voice can seem more hoarse or with increased vocal fry when close to the microphone.}

\subsection{Shout Features}
\begin{figure}
  \centering
  \includegraphics[width=0.6\textwidth]{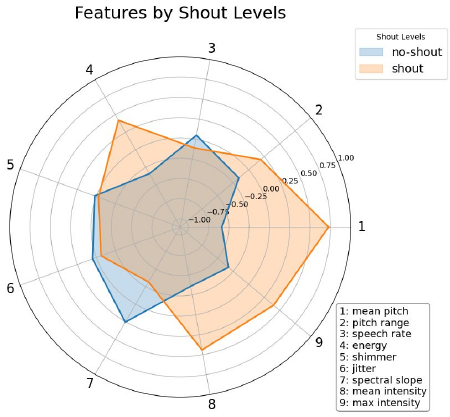}
  \caption{Extracted vocal features for each shout level (normalized).}
  \label{shoutall}
\end{figure}

We inspected how audio features of shouted and non-shouted audio differ. Looking at the entire dataset (Figure \ref{shoutall}), we found, as expected, that shouted speech has higher mean and max intensity, as well as a higher pitch, which are all indicative of Lombard speech \cite{lombard_origin}. \textbf{Surprisingly, pitch \emph{range} is also slightly high, which is not often the case for Lombard speech, which tends to have reduced pitch range when creating a higher pitched pressed voice. It is possible that the increased pitch range is a result of the emotional expressivity of the speech}. Non-shouted speech shows a higher spectral slope, which follows the results in \cite{readroom}.

\subsection{Emotion Features}
Lastly, we investigated how audio features differed for different intended emotions. We divided our plots into typically high and low arousal emotions \cite{RUSSELL1977273} for ease of viewing.

\begin{figure}
  \centering
  \includegraphics[width=\textwidth]{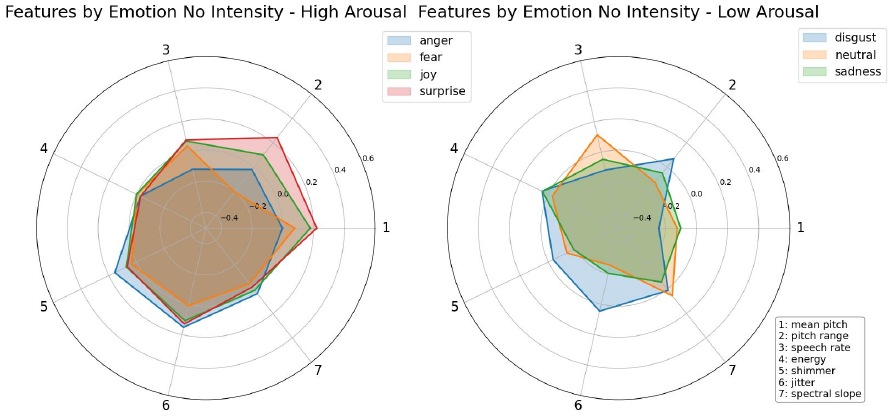}
  \caption{Extracted vocal features for each emotion (normalized) for high (left) and low (right) arousal emotions.}
  \label{emoall}
\end{figure}

When looking at the full dataset in Figure \ref{emoall}, it appears that emotions have less striking differences in features when compared to shout and distances.  It appears the difference between high and low arousal emotions are stronger than that between the emotions within these grouping. Low arousal emotions have a lower pitch, shimmer, and jitter than the high arousal emotions, with the exception of disgust that has a higher shimmer and jitter than the other low arousal emotions. Neutral also has a higher speaking rate than the other low arousal emotions. The high arousal emotions seem to be primarily differentiated by pitch range that is highest for surprise, then joy, followed by anger, and lastly fear. We similarly see joy and surprise having a higher mean pitch than fear and anger. Lastly, anger shows a lower speech rate than the other high arousal emotions.

\textbf{Overall, we see that differences in intensity are primarily driven by both shout and distance levels, however, more strongly by shout. This suggests that the combination of shout and distance can cause confusion for models relying on this feature for classification, such as in SER}. Shimmer and jitter are driven by the phone location, and pitch and pitch range by the shout level. However, the emotion can also play a part in the pitch range. In general, the emotions play a weaker role in determining vocal characteristics than the distance and shouting, but differences can be seen between low and high arousal emotions.

\section{Benchmarks}
\subsection{Automatic Speech Recognition}

We suspected that our dataset would pose a challenge, and therefore would be an important benchmark for real-world, performance of ASR models, especially for models deployed on smart phone devices.  We tested 5 ASR models: Whisper (medium.en) \cite{radford2022robustspeechrecognitionlargescale}\footnote{https://huggingface.co/openai/whisper-medium.en}, Whisper (turbo)\footnote{https://huggingface.co/openai/whisper-large-v3-turbo}, NVIDIA NeMo\footnote{https://github.com/NVIDIA/NeMo} Quartznet \cite{quartz} implementation, NVIDIA NeMo Fastconformer Transducer \cite{fastconf} implementation\footnote{https://catalog.ngc.nvidia.com/orgs/nvidia/teams/nemo/models/\\stt\_en\_fastconformer\_transducer\_large}, and Wav2Vec2-Base-960h \cite{w2v2}\footnote{https://huggingface.co/facebook/wav2vec2-base-960h}. The models' details can be found in Table \ref{asr_models}.

\begin{table}[ht]
  \centering
  \footnotesize
  \renewcommand{\arraystretch}{1.2} 
  \begin{tabular}{
    |>{\raggedright\arraybackslash}p{2.6cm} 
    |>{\raggedright\arraybackslash}p{3cm} 
    |>{\raggedright\arraybackslash}p{1.8cm} 
    |>{\raggedright\arraybackslash}p{2.9cm} 
    |>{\raggedright\arraybackslash}p{1.2cm} |
  }
    \hline
    \textbf{Model} & \textbf{Architecture} & \textbf{\# Param.} & \textbf{Dataset} & \textbf{\# h} \\
    \hline
    Whisper-medium.en & sequence-to-sequence & 769 M & OpenAI custom\textsuperscript{**} & 680,000 \\
    
    Whisper-turbo & sequence-to-sequence\textsuperscript{*} & 798 M & OpenAI custom\textsuperscript{**} & 680,000 \\
    
    NeMo Quartznet & CNN\textsuperscript{\dag} with residual connections & 18.9 M & LibriSpeech\textsuperscript{\ddag}, Common Voice\textsuperscript{\ddag} & 3,500 \\
    
    NeMo FastConformer Transducer & autoregressive conformer & 114 M & NeMo ASRSet & 24,500 \\
    
    Wav2Vec2-Base-960h & CNN-based encoder transformer & 94.4 M & LibriSpeech & 960 \\
    \hline
  \end{tabular}
  
  \caption{ASR model summaries including general model architecture, number of parameters, pretraining dataset, and total hours of pretraining.}
  \label{asr_models}
  
  \vspace{0.5em}
  \parbox{0.9\linewidth}{
    \scriptsize
    \textsuperscript{*} Same as other Whisper models but with 4 instead of 32 decoding layers.\\
    \textsuperscript{**} Multilingual.\\
    \textsuperscript{\dag} 1D time-channel separable convolutions.\\
    \textsuperscript{\ddag} Includes 10\% speed perturbations; English validated set.
  }
\end{table}

\begin{table}[]
  \centering
  \footnotesize
  \begin{tabular}{| l | l | l | l |}
    \hline
    {\textbf{Model}} & {\textbf{WER}} $\downarrow$ & {\textbf{CER}} $\downarrow$ & {\textbf{{PER}}} $\downarrow$\\
    \hline
    Whisper - medium.en  & 31.15\% & 15.13\%  & 14.23\% \\
    Whisper - turbo & \textbf{29.55}\% & \textbf{13.72}\% & \textbf{12.67}\%\\
    NeMo Quartznet &  65.16\% & 32.20\% & 32.80\%\\
    NeMo Fastconformer Transducer& 40.61\% & 22.33\% & 21.49\%\\
    Wav2Vec2-Base-960h & 76.21\% & 35.28\% & 36.52\%\\
    \hline
  \end{tabular}
    \caption{ASR benchmark results over all data.}
    \label{ASR_all}
\end{table}

The results over the entire dataset can be seen in Table \ref{ASR_all} where we report word error rate (WER), the percentage of transcribed words correctly matching the ground truth transcription, character error rate (CER), the percentage of transcribed characters correctly matching the ground truth transcription, and the phoneme error rate (PER), the percentage of phonemized elements of the transcription correctly matching the ground truth transcription (both transcriptions phonemized with espeak-ng\footnote{https://github.com/espeak-ng/espeak-ng}) for each model. We see both the Whisper models showing the best performance, followed by the Fastconformer Transducer. Yet, all the models perform worse (between 29.55 and 76.21\% WER) than the usual reported WERs, which are often below 10\%. In particular, Whisper medium.en is reported to have a WER of 27\% on the CHiME-6 dataset \cite{radford2022robustspeechrecognitionlargescale, watanabe2020chime} that is similarly used for distanced ASR benchmarking.

\begin{table}[ht]
\centering
\begin{minipage}[t]{0.49\textwidth}
  \resizebox{\textwidth}{!}{
  \begin{tabular}{| l | l | l | l |}
    \hline
    {\textbf{Subset}} & {\textbf{WER}} $\downarrow$ & {\textbf{CER}} $\downarrow$ & {\textbf{PER}} $\downarrow$\\
    \hline
    Disgust  & 23.96\% & \textbf{11.26}\%  & 10.92\% \\
    Joy & 31.87\% & 15.93\% & 14.40\%\\
    Neutral &  \textbf{23.20}\% & 11.35\% & \textbf{10.40}\% \\
    Anger & 29.3\% & 13.48\% & 12.49\%\\
    Fear & 35.73\% & 17.25\% & 16.42\%\\
    Surprise & 35.58\% & 17.50\% & 16.75\%\\
    Sadness & 39.42\% & 19.82\% & 18.91\%\\
    \hline
    Near Body & \textbf{24.93}\% & \textbf{11.78}\% & \textbf{11.27}\%\\
    1-2m Away & 27.00\% & 12.96\% & 11.77\%\\
    Other side of room & 35.02\% & 16.26\% & 15.59\%\\
    Outside of room & 45.10\% & 23.64\% & 21.99\%\\
    \hline
    Shout & 39.83\% & 19.43\% & 17.72\%\\
    No Shout & \textbf{24.17}\% & \textbf{11.67}\% & \textbf{10.95}\%\\
    \hline
  \end{tabular}}
    \caption{ASR benchmark results for Whisper medium.en.}
    \label{whisperm}
\end{minipage}
\hfill
\begin{minipage}[t]{0.49\textwidth}
  \centering
\resizebox{\textwidth}{!}{
  \begin{tabular}{| l | l | l | l |}
    \hline
    {\textbf{Subset}} & {\textbf{WER}} $\downarrow$ & {\textbf{CER}} $\downarrow$ & {\textbf{PER}} $\downarrow$\\
    \hline
    Disgust  & 23.29\% & 10.21\%  & 10.04\% \\
    Joy & 30.17\% & 14.23\% & 12.74\%\\
    Neutral &  \textbf{22.31}\% & \textbf{9.86}\% & \textbf{9.15}\%\\
    Anger & 25.98\% & 11.34\% & 10.85\%\\
    Fear & 34.06\% & 15.28\% & 14.70\%\\
    Surprise & 32.94\% & 16.19\% & 17.9\%\\
    Sadness & 39.10\% & 19.65\% & 13.6\%\\
    \hline
    Near Body & \textbf{22.44}\% & \textbf{9.64}\% & \textbf{9.35}\%\\
    1-2m Away & 26.75\% & 11.93\% & 12.55\%\\
    Other side of room & 32.08\% & 14.43\% & 13.89\%\\
    Outside of room & 45.32\% & 23.80\% & 18.71\%\\
    \hline
    Shout & 37.39\% & 17.94\% & 15.99\%\\
    No Shout & \textbf{23.29}\% & \textbf{10.32}\% & \textbf{10.01}\%\\
    \hline
  \end{tabular}}
    \caption{ASR benchmark results for Whisper turbo.}
    \label{whispert}
\end{minipage}
\end{table}

\begin{figure}[h]
  \centering
  \includegraphics[width=\textwidth]{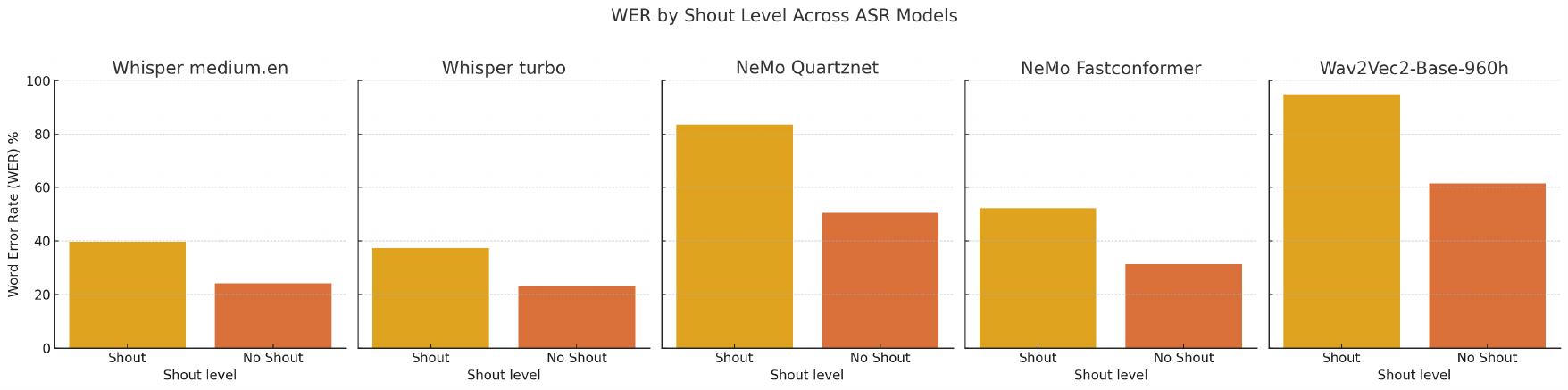}
  \caption{WER per shout level across the benchmarked ASR models.}
  \label{images-distance}
\end{figure}

\begin{figure}[h]
  \centering
  \includegraphics[width=\textwidth]{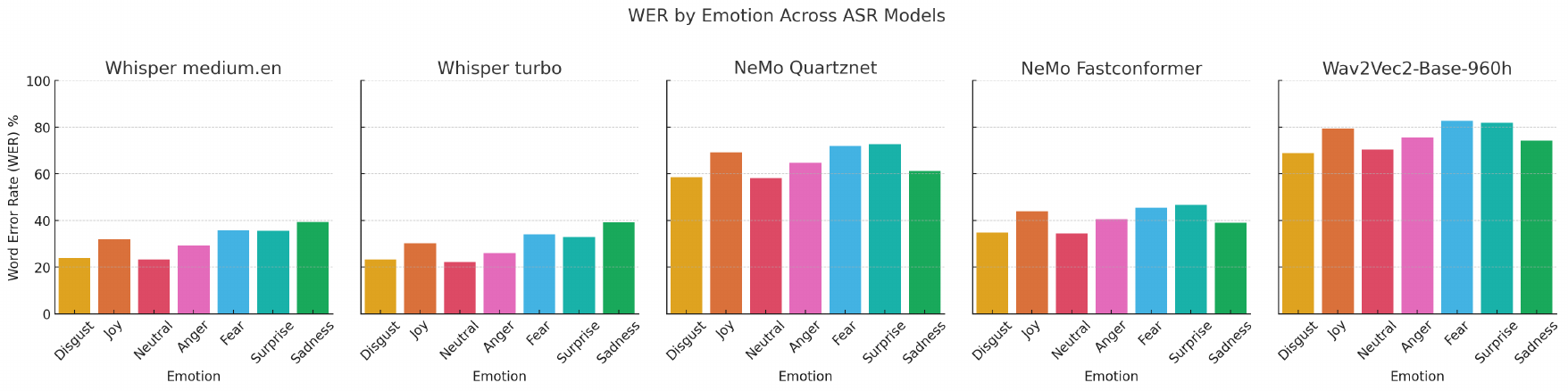}
  \caption{WER per emotion across the benchmarked ASR models.}
  \label{images-emotion}
\end{figure}

We further investigated the performance of these ASR models to better understand how performance differs over different distances and for different emotions and shout levels. The results can be seen in Tables \ref{whisperm} - \ref{asrw2v2} and in Figures \ref{images-distance} - \ref{images-emotion}. For all models, we see a degradation of WER performance as the phone moves further away from the actor. No particular model appears to be markedly more robust to this difficulty than other, however, NeMo Fastconformer Transducer did show the lowest reduction in WER from 33.68 - 53.44\%: Whisper (medium.en) showed 21\% degradation from near the body to in the other room, Whisper (turbo) 23\%, NeMo Quartznet 27\%, NeMo Fastconformer Transducer 20\%, and Wav2Vec2-Base-960h 28\%. 

\begin{figure}[h]
  \centering
  \includegraphics[width=\textwidth]{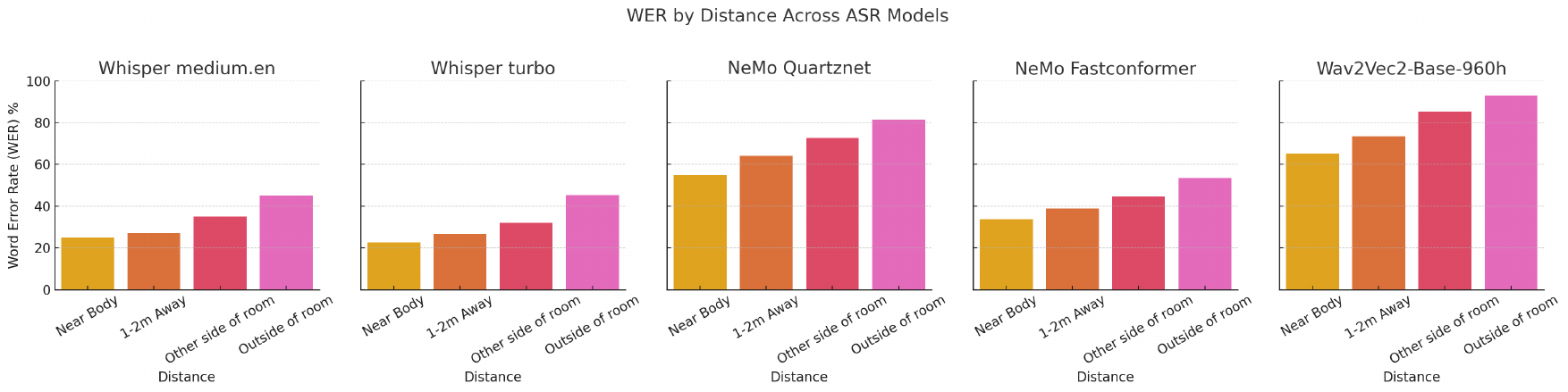}
  \caption{WER per distance across the benchmarked ASR models.}
  \label{images-shout}
\end{figure}

\begin{table}[H]
\centering
\begin{minipage}[t]{0.48\textwidth}
  \resizebox{\textwidth}{!}{
  \begin{tabular}{| l | l | l | l |}
    \hline
    {\textbf{Subset}} & {\textbf{WER}} $\downarrow$ & {\textbf{CER}} $\downarrow$ & {\textbf{PER}} $\downarrow$\\
    \hline
    Disgust  & 58.42\% & 28.19\%  & 28.75\% \\
    Joy & 69.07\% & 35.39\% & 35.24\%\\
    Neutral &  \textbf{58.21}\% & \textbf{27.10}\% & \textbf{27.81\%}\\
    Anger & 64.77\% & 30.31\% & 30.47\%\\
    Fear & 71.89\% & 36.84\% & 37.98\%\\
    Surprise & 72.65\% & 36.45\% & 37.69\%\\
    Sadness & 61.22\% & 31.10\% & 31.76\%\\
    \hline
    Near Body & \textbf{54.72}\% & \textbf{25.30}\% & \textbf{26.07}\%\\
    1-2m Away & 64.00\% & 30.08\% & 30.77\%\\
    Other side of room & 72.54\% & 35.79\% & 36.64\%\\
    Outside of room & 81.27\% & 45.83\% & 45.60\%\\
    \hline
    Shout & 83.45\% & 45.16\% & 45.81\%\\
    No Shout & \textbf{50.54}\% & \textbf{21.77}\% & \textbf{22.33}\%\\
    \hline
  \end{tabular}}
    \caption{ASR benchmark results for NeMo Quartznet.}
    \label{asrquartz}
\end{minipage}
\hfill
\begin{minipage}[t]{0.48\textwidth}
  \centering
\resizebox{\textwidth}{!}{
  \begin{tabular}{| l | l | l | l |}
    \hline
    {\textbf{Subset}} & {\textbf{WER}} $\downarrow$ & {\textbf{CER}} $\downarrow$ & {\textbf{PER}} $\downarrow$\\
    \hline
    Disgust  & 34.71\% & 17.91\%  & \textbf{17.23}\% \\
    Joy & 43.98\% & 24.83\% & 23.32\%\\
    Neutral &  \textbf{34.41}\% & \textbf{17.74}\% & 17.42\%\\
    Anger & 40.45\% & 20.29\% & 19.41\%\\
    Fear & 45.35\% & 26.14\% & 25.21\%\\
    Surprise & 46.69\% & 26.18\% & 37.69\%\\
    Sadness & 39.06\% & 23.38\% & 22.56\%\\
    \hline
    Near Body & \textbf{33.68}\% & \textbf{16.65}\% & \textbf{16.13}\%\\
    1-2m Away & 38.92\% & 20.57\% & 19.67\%\\
    Other side of room & 44.44\% & 24.20\% & 23.19\%\\
    Outside of room & 53.44\% & 34.67\% & 33.32\%\\
    \hline
    Shout & 52.28\% & 31.46\% & 30.51\%\\
    No Shout & \textbf{31.35}\% & \textbf{14.99}\% & \textbf{14.25}\%\\
    \hline
  \end{tabular}}
    \caption{ASR benchmark results for NeMo Fastconformer Transducer.}
    \label{asrfast}
\end{minipage}
\end{table}

\begin{table}[H]
  \centering
  \footnotesize
  \begin{tabular}{| p{0.21\textwidth} | p{0.1\textwidth}| p{0.09\textwidth} | p{0.1\textwidth}|}
    \hline
    {\textbf{Subset}} & {\textbf{WER}} $\downarrow$ & {\textbf{CER}} $\downarrow$ & {\textbf{PER}} $\downarrow$\\
    \hline
    Disgust  & \textbf{68.73}\% & \textbf{30.82}\%  & \textbf{32.20}\% \\
    Joy & 79.26\% & 37.65\% & 38.09\%\\
    Neutral &  70.33\% & 31.53\% & 32.70\%\\
    Anger & 75.50\% & 33.94\% & 34.75\%\\
    Fear & 82.62\% & 38.94\% & 40.80\%\\
    Surprise & 81.82\% & 38.78\% & 39.97\%\\
    Sadness & 74.18\% & 35.32\% & 36.78\%\\
    \hline
    Near Body & \textbf{65.29}\% & \textbf{27.88}\% & \textbf{29.33}\%\\
    1-2m Away & 73.33\% & 33.40\% & 34.25\%\\
    Other side of room & 85.37\% & 40.30\% & 42.10\%\\
    Outside of room & 92.92\% & 48.26\% & 48.69\%\\
    \hline
    Shout & 94.75\% & 46.47\% & 48.11\%\\
    No Shout & \textbf{61.58}\% & \textbf{26.32}\% & \textbf{27.30}\%\\
    \hline
  \end{tabular}
    \caption{ASR benchmark results for Wav2Vec2-Base-960h.}
    \label{asrw2v2}
\end{table}

All models struggled to transcribe shouted speech. The Whisper models were the most robust with a 15\% (medium.en) and 14\% (turbo) increase in WER. NeMo Quartznet and Wav2Vec2-Base-960h stuggled the most with shouted speech with a 33\% increase in WER for shouted speech.

For all models there was a difference in performance across the emotions, though less severe than for distance. Sentences produced with intended neutral and disgusted voices were the easiest emotions to transcribe across all models, while high arousal emotions like joy, fear, and surprise tended to be difficult to transcribe. Outliers include the Whisper models, that tended to struggle the most with sadness. It is not particularly surprising that the models show the highest performance on neutral speech, as this is often what they are trained and benchmarked on. Once again, the NeMo Fastconformer Transducer appears to be slightly more robust to differences in performance in the WER across the emotions. Unlike other emotional speech datasets, BERSt uses the same phrases for all emotions, and therefore, unlike the conclusions in \cite{li2023asremotionalspeechwordlevel}, we cannot propose that the discrepancies in WER are related to the use of emotionally specific wording. Instead, our results derive purely from the differences in vocal features between emotions.

\begin{table}[]
  \centering
  \footnotesize
  \begin{tabular}{| l | l | l | l |}
    \hline
    {\textbf{Model}} & {\textbf{WER}} $\downarrow$ & {\textbf{CER}} $\downarrow$ & {\textbf{PER}} $\downarrow$\\
    \hline
    Whisper - medium.en  & \textbf{17.27}\% & 7.81\%  & 7.80\% \\
    Whisper - turbo & 17.93\% & \textbf{7.28}\% & \textbf{7.30}\%\\
    NeMo Quartznet &  39.49\% & 15.24\% & 15.77\%\\
    NeMo Fastconformer Transducer& 24.96\% & 10.72\% & 10.13\%\\
    Wav2Vec2-Base-960h & 49.65\% & 18.94\% & 19.90\%\\
    \hline
  \end{tabular}
    \caption{ASR benchmark results for neutral, non-shouted data, recorded next to the body.}
    \label{ASR_norm}
\end{table}

We then investigated how the models performed on non-shouted data, that is neutral and next to the body, as this is expected to be similar to most other ASR datasets. The results can be seen in Table \ref{ASR_norm}. All the models see a drastic increase in performance when the most difficult data is removed, yet, is still lower than expected for these models. One reason may be due to the lack of linguistic context in these phrases. It may also stem from the obstructions applied to the phone next to the body, such as placing the phone in a bag or a pocket for some of the recordings.

\subsection{Speech Emotion Recognition}
Although the emotions expressed by professional actors in our data have yet to be validated by perception based annotations, we provide an initial benchmark to display another use case for the BERSt dataset. Therefore, these results must be viewed through the lens of the intended emotions produced by the professional actors, and not necessarily those perceived by listeners. The results are presented as confusion matrices, i.e. counts of the predicted label (x-axis) given the ground truth label (y-axis). Moreover, the unweighted (averaged over all samples) and weighted (weights assigned per class in case of class imbalance before averaging) accuracies are provided for all of the models.

SER remains a difficult problem and overall sees lower performance than ASR tasks. One problem is that there remains no ``gold standard" for classifying emotions and many SER models are only trained to classify 4 emotions: neutral, anger, joy and sadness. However, there has been a recent surge of models that can provide a continuous classification of emotions using valence, arousal, and dominance scores. Given that our data has 7 classes, we modified all of the models to be benchmarked through some form of fine-tuning. We tested three speech emotion recognition models: a Wav2Vec2 based classifier \cite{w2v2emo} implemented in SpeechBrain \cite{speechbrain}, DAWN from audEERING \cite{dawn}, which uses a similar backbone but is fine-tuned on MSP-Podcast \cite{Lotfian_2019_3} for continuous valence, arousal, and dominance predictions, and Wav2Small \cite{kounadisbastian2024wav2smalldistillingwav2vec272k} that combines the DAWN model and Odyssey WavLM model \cite{Goncalves_2024} (only the teacher model is publicly available, so no distilling is applied). The models' details can be found in Table \ref{ser_models}.

\begin{table}[]
  \centering
  \footnotesize
  \begin{tabular}{
      | >{\raggedright\arraybackslash}p{2.7cm}
    | >{\raggedright\arraybackslash}p{4cm}
    | >{\raggedright\arraybackslash}p{1.8cm}
    | >{\raggedright\arraybackslash}p{2.2cm}
    | >{\raggedright\arraybackslash}l |}
    \hline
    {\textbf{Model}} & {\textbf{Arch.}} & {\textbf{\# Param.}} & {\textbf{Dataset}} &
    {\textbf{\# h}}\\
    \hline
    SpeechBrain Wav2Vec2  & \raggedright Classification layer on Wav2Vec2* \arraybackslash & 165 M & IEMOCAP-4 Fine-tune: IEMOCAP-7& 12\\
    DAWN-hidden-SVM & \raggedright Final layer for VAD on Wav2Vec2 \arraybackslash & 165 M & MSP-Podcast Fine-tune: BERSt & 237\\
    Wav2Small-VAD-SVM* & DAWN model fused with Odyssey WavLM** & 483.9 M & MSP-Podcast Fine-Tune: BERSt & 237\\
    \hline
  \end{tabular}
    \caption{SER model summaries including the general model architecture, the number of parameters, the pretraining dataset and the number of hours of pretraining data.}
    \label{ser_models}
\parbox{5in}{
\scriptsize{*as in Table \ref{asr_models}\\
**transformer encoder
}}
\end{table}

\begin{figure}[ht]
    \centering
    \begin{minipage}{0.44\textwidth}
        \centering
        \includegraphics[width=\linewidth]{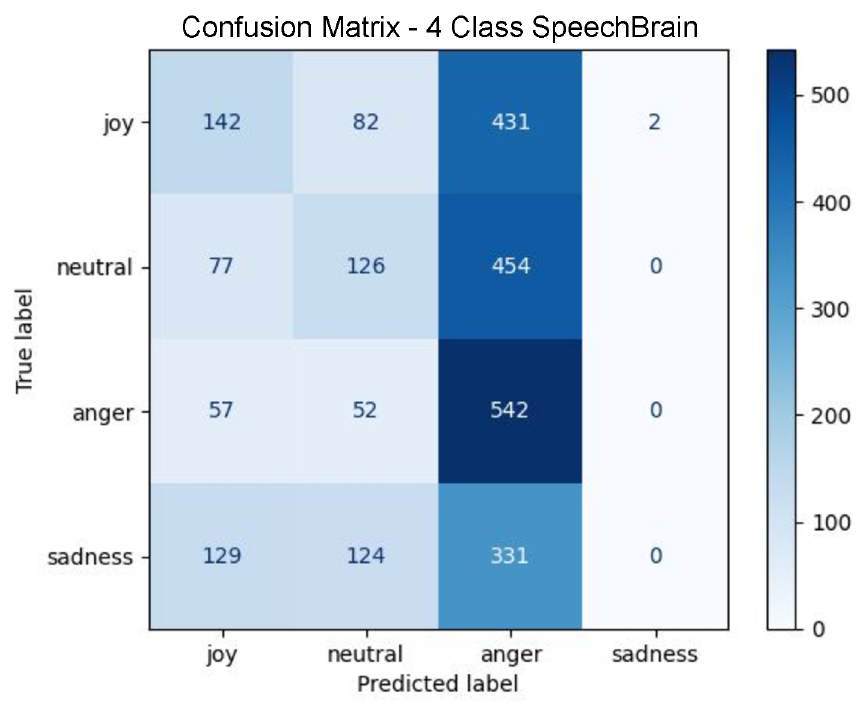}
        \caption{SER confusion matrix of out-of-box prediction on the 4-class \\subset of BERSt using the SpeechBrain model.}
        \label{4-speech-out-of-box}
    \end{minipage}%
    \hfill
    \begin{minipage}{0.54\textwidth}
        \centering
        \includegraphics[width=\linewidth]{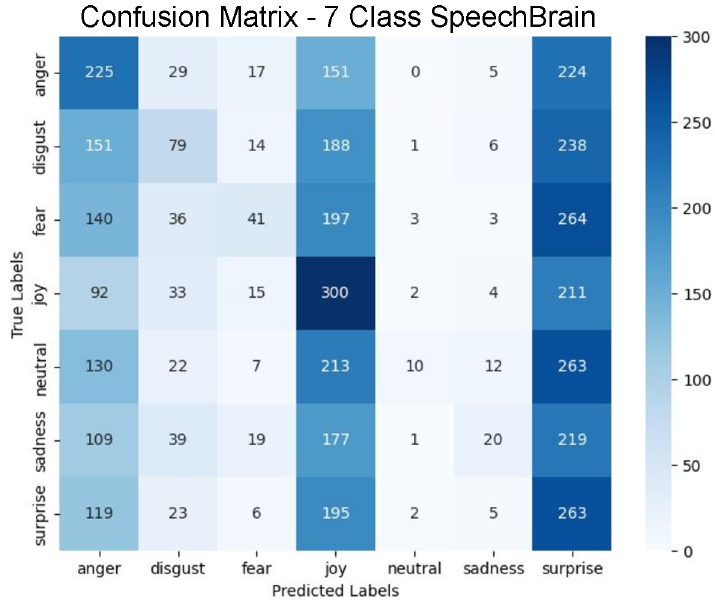}
        \caption{SER confusion matrix of IEMOCAP-7 fine-tuned 7-class classification with the \\ SpeechBrain model.}
        \label{7-speech-finetuned}
    \end{minipage}
\end{figure}

These models use the same backbone, one building upon the other. As such, we present the results from the base model and explore if the addition of training data and modifications on the final layer (DAWN) or the fusion of a secondary model (Wav2Small) make the model more robust to our difficult data. We do our best to ensure minimal fine tuning on the model to test out-of-box performance similar to the ASR experiments. However, because the models are either only for a) 4 class classification or b) for continuous prediction of valence, arousal, and dominance, a slight fine-tuning of the final layers (SpeechBrain) of the model, or the addition of a classifier (DAWN and Wav2Small) was needed.

\begin{figure}[H]
    \centering
    \begin{minipage}{0.54\textwidth}
        \centering
        \includegraphics[width=\linewidth]{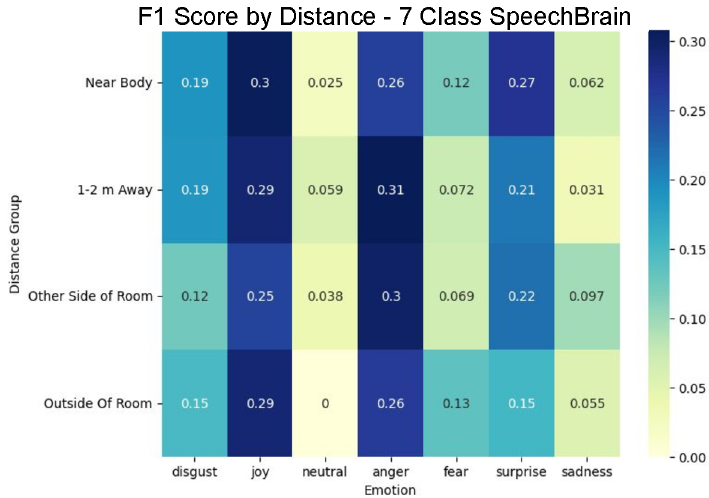}
        \caption{SER F1 scores per distance and emotion for BERSt fine-tuned 7 class classification with the SpeechBrain model.}
        \label{distance-sb}
    \end{minipage}%
    \hfill
    \begin{minipage}{0.44\textwidth}
        \centering
        \includegraphics[width=\linewidth]{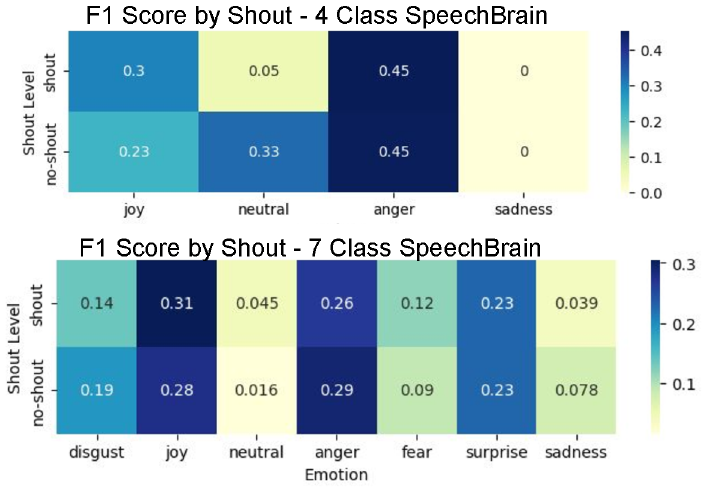}
        \caption{SER F1 scores per shout level and emotion for out-of-box 4 class classification (top) and BERSt \\ fine-tuned 7 class classification (bottom) with the SpeechBrain model.}
        \label{shout-sb}
    \end{minipage}
\end{figure}

\textbf{Out-of-box SpeechBrain.} For the SpeechBrain model, we first attempted an out-of-box classification on only 4 emotions, as this was the only model that we could test out-of-box without adding some form of fine tuning. We created a subset of the BERSt dataset to keep only the audio samples of the emotions joy, neutral, anger, and sadness. It can be seen that the majority of the data is misclassified as anger (Figure \ref{4-speech-out-of-box}) and only two utterances are classified as sadness (both of which were acted as joy). This may be due to the high intensity and arousal of the shouting in the dataset. This can be seen particularly strongly in the neutral class (Fig. \ref{shout-sb} (top)) where the F1-score for non-shouted speech is much higher than for shouted speech in this, classically, low arousal emotion. Sadness, however, has too few classifications to understand why so many misclassifications are occurring, yet, during cleaning it was noted by the annotators that the actors were using a high arousal ``sobbing" form of sadness that was easily confused with joy and anger. Likely, the poor results in the sadness classification relates to differences in sub-emotions of sadness (depression vs. grief) and a specific interpretation of this emotion by the actors for the training dataset than for those for BERSt. Specifically, as the actors were shouting during much of their recording sessions, the high-arousal form of grief may be favored in the data. These subtle differences in emotion classification, both for machine learning models and for humans are so complex and many researchers have moved towards the valence-arousal-dominance models seen with the DAWN models, as this classification scheme is able to capture more subtle differences within emotions.

\textbf{IEMOCAP-7 Fine-tuned SpeechBrain.} We then used the fine-tuning approach provided by SpeechBrain\footnote{https://github.com/speechbrain/speechbrain/tree/develop/\\recipes/IEMOCAP} for 25 epochs on the full, 7-class IEMOCAP dataset to allow for 7 class classification, changing the output classification layer to size 7, rather than 4. The IEMOCAP-7 dataset contains fewer samples of disgust, fear, and surprise than for the 4 emotions included in IEMOCAP-4. This may have an impact on the results for the emotion classes. The confusion matrix for the results of the full BERSt dataset can be seen in Figure \ref{7-speech-finetuned}. Although we do see some more data being classified into the low arousal emotions, we still see a strong bias towards all the data being classified into high arousal emotions: joy, anger, and surprise.

We also explored the impact of distance (Figure \ref{distance-sb}) and shout level (Figure \ref{shout-sb})  on the classification of both the 4 and 7 class models by plotting the F1-score of the classified emotions (x-axis) by the distance and shout levels (y-axis). The contribution of these features to the confusion of emotions, however, seems much smaller than for ASR tasks, possibly due to the fine tuning. To confirm this, in Figure \ref{shout-sb}(top) we explored the shout level results with the 4 class out-of-box model and it can be seen that without fine tuning the low arousal, neutral does see performance degradation for shouted speech. 

\textbf{VAD-to-classifier DAWN model}. The DAWN model provides a fine-tuning method to convert the probability scores of three emotional measures valence, arousal, and dominance to classes using an SVM\footnote{url{https://github.com/audeering/w2v2-how-to}}. We tried two SVM approaches: 1) an SVM on the valence, arousal, and dominance results, and 2) an SVM on the 1024 hidden feature layer before the valence, arousal, and dominance prediction layer. The SVM on the hidden layer showed the best performance, and as such these are the results presented.

\begin{figure}[ht]
    \centering
    \begin{minipage}{0.49\textwidth}
        \centering
        \includegraphics[width=\linewidth]{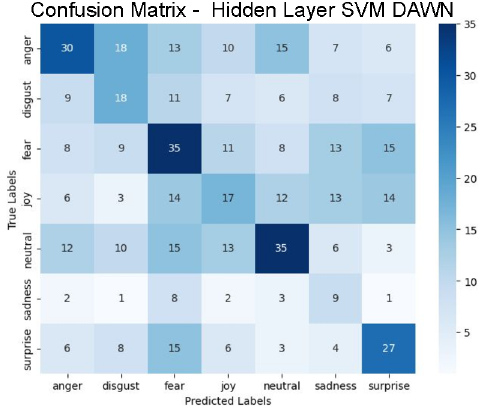}
        \caption{Confusion Matrix of DAWN model results with an SVM on the hidden layers.}
        \label{conf-dawn}
    \end{minipage}%
    \hfill
    \begin{minipage}{0.49\textwidth}
        \centering
        \includegraphics[width=\linewidth]{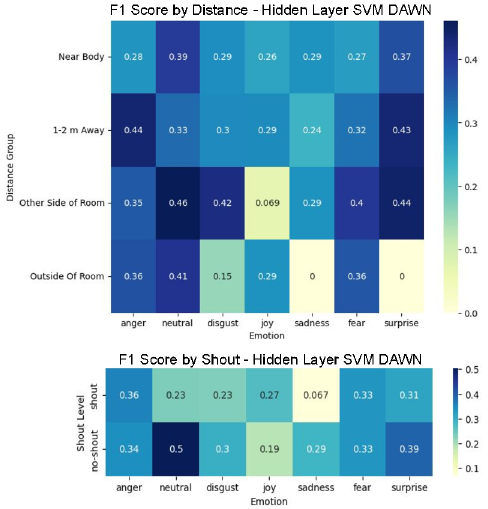}
        \caption{SER F1 scores per distance level (top) and shout level (bottom) for the DAWN model with an SVM on the hidden layers.}
        \label{heat-dawn}
    \end{minipage}
\end{figure}

The classification results on the test set can be seen in Figure \ref{conf-dawn}. We see the model struggles to classify sadness, confusing it with fear and joy. Other emotions appear to have confusion with several other emotions, both low and high arousal. In Figure \ref{heat-dawn} we see the F1 score results for each emotion over distance and shout levels. We see that there is not a strong pattern in degradation of the results over distance (top), with the exception being outside of the room. For the shout levels (bottom) we see that the typically low arousal emotions of neutral and sadness show reduced performance when shouted.

\textbf{VAD-to-classifier Wav2Small}. Lastly, the Wav2Small teacher model similarly provides a continuous valence, arousal, and dominance result. We once again tested the two SVM approaches to convert the results for classification. In the end, we used an SVM on the valence, arousal, and dominance results as this performed better than classification on the hidden layers for this fused model.

\begin{figure}[H]
    \centering
    \begin{minipage}{0.52\textwidth}
        \centering
        \includegraphics[width=\linewidth]{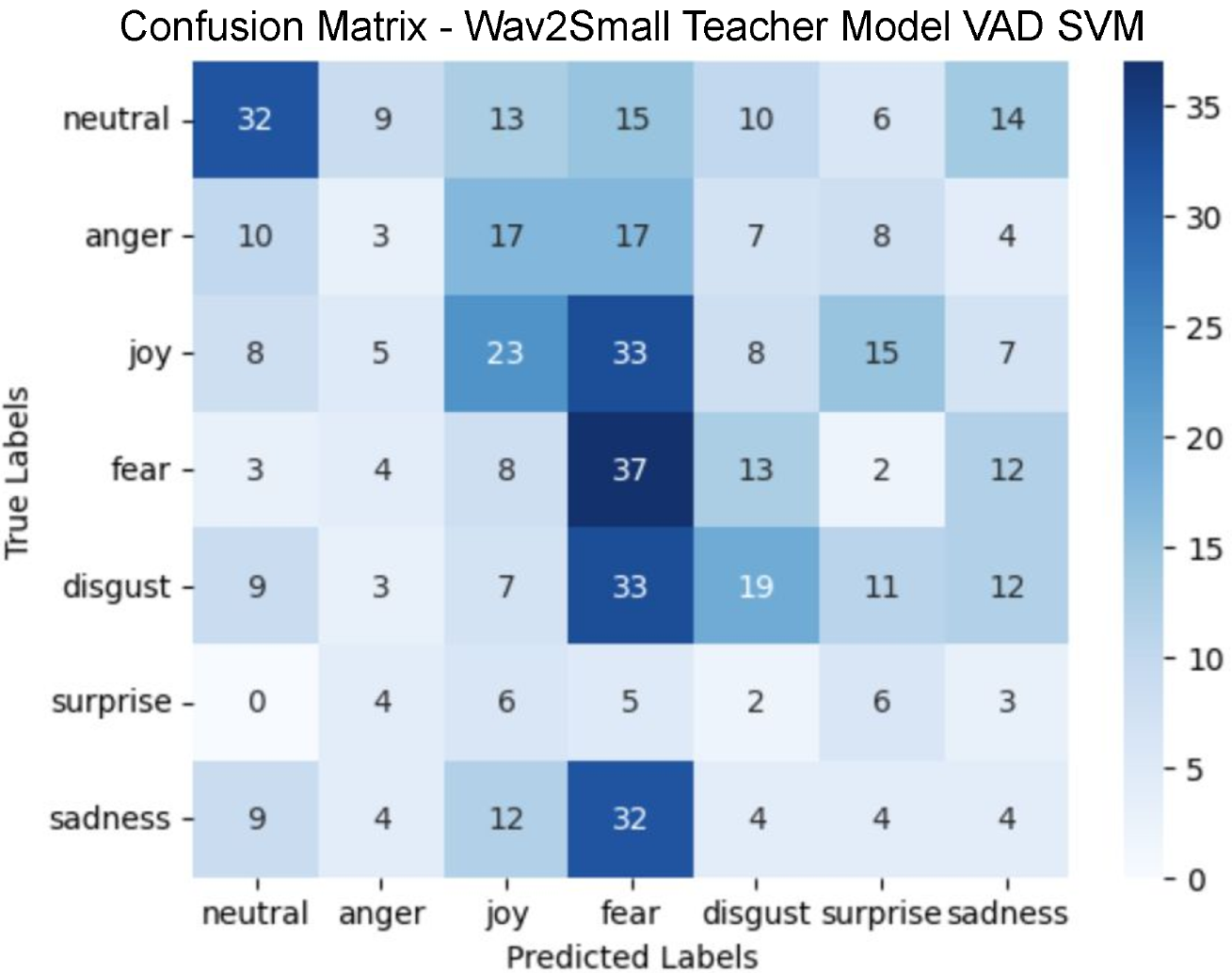}
        \caption{Confusion Matrix of the Wav2Small teacher model with an SVM on the VAD outputs.}
        \label{conf-w2small}
    \end{minipage}%
    \hfill
    \begin{minipage}{0.46\textwidth}
        \centering
        \includegraphics[width=\linewidth]{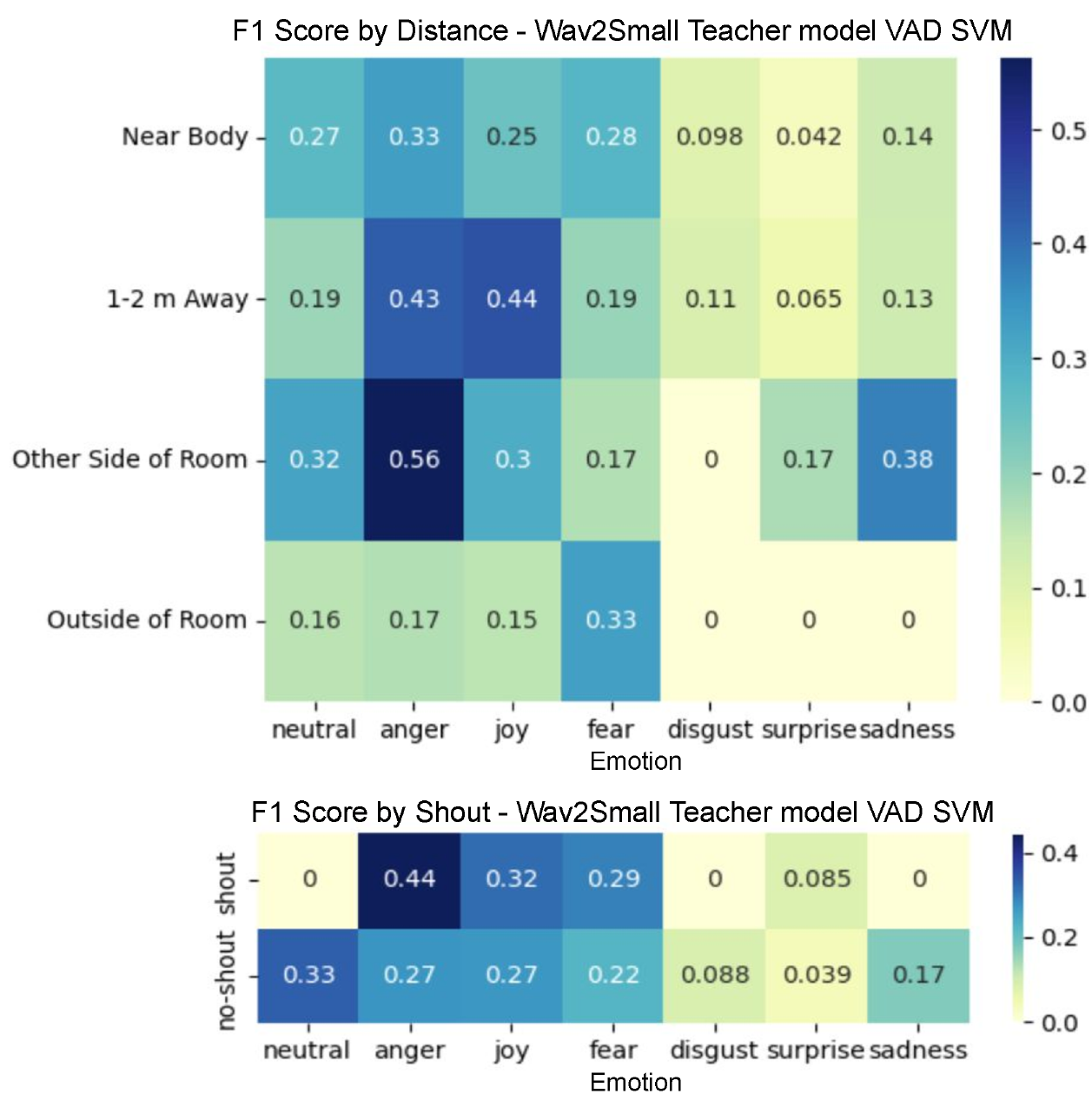}
        \caption{SER F1 scores per distance level (top) and shout level (bottom) for the Wav2Small teacher model with an SVM on the VAD outputs.}
        \label{heat-w2small}
    \end{minipage}
\end{figure}

The classification results on test set can be seen in Figure \ref{conf-w2small}. The model over-predicts fear and struggles to predict anger. In Figure \ref{heat-w2small}, we see the F1 score results for each emotion over distance and shout levels. Once again we only seem to have a strong reduction in F1 score for the out of the room distance (top), except for fear. However, for shouts (bottom) all the low arousal emotions see a drastic reduction in F1 score, while most of the high arousal emotions actually see improvement. This could suggest that this model is particularly sensitive to intensity levels.

The results for weighted and unweighted accuracy for each of the three models can be seen in Table \ref{SER_all}. DAWN, showing the best performance, and Wav2Small showing the worst performance. The introduction of the Odyssey model to the DAWN appears to have a noteworthy impact on lowering the performance with the BERSt dataset. Lastly, as suggested in \cite{li2023asremotionalspeechwordlevel}, the reduced WER we see in ASR models may suggest that performance of SER will be lower for our dataset.

Overall, all of these models struggle with the complexity of the BERSt dataset. When the models have never been fine-tuned on BERSt (Speechbrain), there is a strong bias towards high arousal emotions. It is known that SER models do not generalize well to new datasets \cite{ibrahim24_interspeech}, which is likely due to the subtle differences in interpretations of emotions and the need for an overall more robust classification system from annotation to training. The DAWN model, initially trained on a continuous rather than classification task, shows improved results, perhaps due to this underlying modeling of VAD or the fine-tuning with BERSt to help the model generalize to a new dataset. Yet, even with fine-tuning, we still see the impact of the phone being at a far distance and the unexpected high arousal in shouting classically low arousal emotions. These distance and shout-based misclassifications are exacerbated by the fusion of the Odessey model in Wav2Small. At the base, all of these models rely on Wav2Vec2, and their struggle in classifying this complex data suggested that further work needs to be done to improve the robusticity to new, more challenging conditions for SER, perhaps through exploring the layers of Wav2Vec2 and understanding which are more robust to changes in shout level and distance as has been done for other acoustic classification tasks \cite{dieck22_interspeech, english24_interspeech, lee24e_interspeech}.

\begin{table}[]
  \centering
  \footnotesize
  \begin{tabular}{| l | l | l |}
    \hline
    {\textbf{Model}} & {\textbf{UA}} $\uparrow$ & {\textbf{WA}} $\uparrow$ \\
    \hline
    SpeechBrain Wav2Vec2  & 20.7\% & 20.8\%  \\
    DAWN-hidden-SVM & \textbf{32.1}\% & \textbf{32.2}\%\\
    Wav2Small-VAD-SVM* &  23.3\% & 22.3\% \\
    \hline
  \end{tabular}
  \begin{minipage}{\linewidth}
  \centering
    \small *Teacher model
  \end{minipage}
    \caption{SER benchmark results for unweighted (UA) and weighted accuracy (WA).}
    \label{SER_all}
\end{table}

\section{Future Directions}
We have provided benchmarks for state-of-the-art models using the BERSt dataset. Further work training models to optimize performance for this dataset remains future work. In addition, we do not provide validated emotional labels, nor continuous valence, arousal, and dominance labels. This remains future work and we welcome researchers to provide any other possible emotion categories specific to their use case. In addition, we provide the first language and current language of the speakers, and we know there are regional accents. This is based on the actors location and our reviews of the audio files, however, it would be useful to, in the future, label the regional accents of the speakers.

\section{Conclusions}
It appears that more work needs to be done to improve the robustness of ASR models to distanced speech, emotional speech, shouted speech, and high surprisal speech. Although the state-of-the art methods we evaluated perform with reported WERs below 10\% on many common test sets, our best performing model, Whisper (turbo), had a WER of 29.55\%. We saw similar problems in the SER models where we observed a maximum of a 32.2\% unweighted average with the DAWN model, which is not an acceptable performance for real-world deployment. Both the SER and ASR models struggle to maintain their performance for shouted speech, and ASR, specially sees performance degredation as the recording device is moved further from the speaker. Even in the case where the phone is near the body and the speech is neutral, ASR methods perform below the expected values. BERSt provides a challenging new benchmark with distanced, shouted, and emotion speech in a variety of home environments, recorded on smartphones to evaluate multiple speech emotion recognition tasks.

\section{Acknowledgments}
This research was supported by Mitacs through the Mitacs E-accelerate program. We acknowledge the support of the Natural Sciences and Engineering Research Council of Canada (NSERC). We thank Maya Becher, and Jackie Lu for their time and help with the dataset, Mohammed Hafsati for his advice and advising and the Rajan Family for their support.
\appendix
\section{Data Collection Application Image}
\label{app1}

\begin{figure}[H]
  \centering
  \includegraphics[width=10cm]{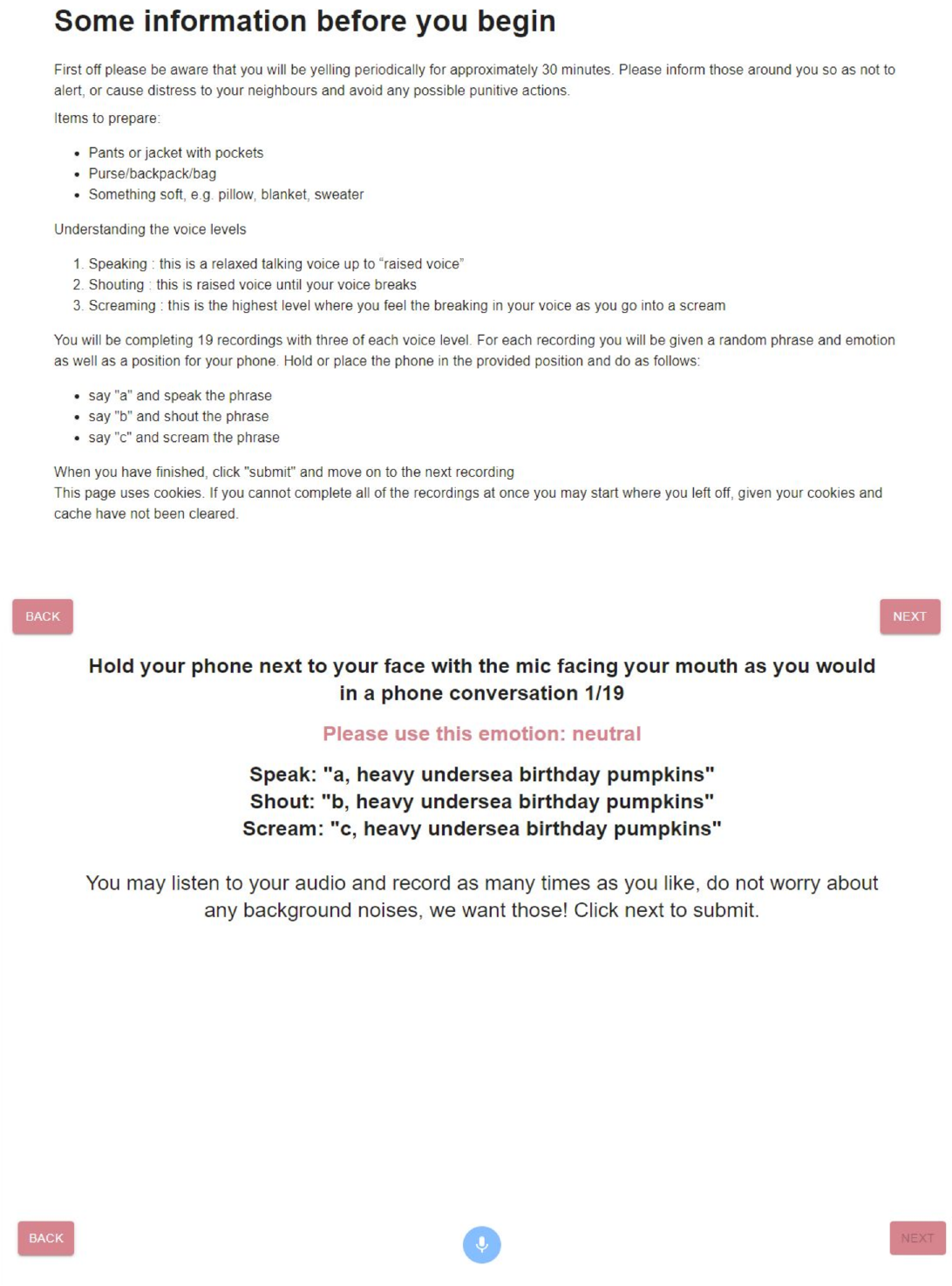}
  \caption{Example pages from the data collection application.}
  \label{app}
\end{figure}

\bibliographystyle{elsarticle-num} 

\end{document}